%% file: main.tex
\documentclass[letterpaper, 10 pt, conference]{ieeeconf}  

\IEEEoverridecommandlockouts                              

\usepackage{graphics} 
\usepackage{epsfig} 
\usepackage{times} 
\usepackage{amsmath} 
\usepackage{amssymb}  
\usepackage{mathtools}

\usepackage[dvipsnames]{xcolor}
\usepackage{booktabs}
\usepackage{graphicx}
\usepackage{subcaption}
\usepackage{cite}

\makeatletter
\let\NAT@parse\undefined
\makeatother

\usepackage{url}
\usepackage{hyperref}

\newcommand\myshade{85}
\colorlet{mylinkcolor}{black}
\colorlet{mycitecolor}{MidnightBlue}
\colorlet{myurlcolor}{RedViolet}

\hypersetup{
	linkcolor  = mylinkcolor!\myshade!black,
	citecolor  = mycitecolor!\myshade!black,
	urlcolor   = myurlcolor!\myshade!black,
	colorlinks = true,
}

\def\shownotes{1}  
\ifnum\shownotes=1
\newcommand{\authnote}[2]{{$\ll$\textsf{\footnotesize #1 notes: #2}$\gg$}}
\else
\newcommand{\authnote}[2]{}
\fi

\DeclareMathOperator*{\argmin}{arg\,min}

\title{\LARGE \bf
	Generalizing Object-Centric Task-Axes Controllers using Keypoints
}

\author{Mohit Sharma$^{1}$ and Oliver Kroemer$^{1}$ \\
	\thanks{$^{1}$Robotics Institute,  Carnegie Mellon University, Pittsburgh, PA, USA}
	\thanks{Correspondence to \texttt{mohits1@cs.cmu.edu}}
}

\begin{document}

	\maketitle
	\thispagestyle{empty}
	\pagestyle{empty}

	\begin{abstract}
		
		To perform manipulation tasks in the real world, robots need to operate on objects with various shapes, sizes and without access to geometric models. It is often unfeasible to train monolithic neural network policies across such large variance in object properties.
		Towards this generalization challenge, we propose to learn 
		modular task policies which compose object-centric task-axes controllers. 
		These task-axes controllers are parameterized by properties
		associated with underlying objects in the scene.
		We infer these controller parameters directly from visual input using multi-view dense correspondence learning.
		Our overall approach provides a simple and yet powerful framework for learning manipulation tasks. 
		We empirically evaluate our approach on 3 different manipulation tasks and show its ability to generalize to large variance in object size, shape and geometry.


	\end{abstract}

	\section{INTRODUCTION}
	\label{sec:intro}
	
	Manipulation tasks in the real world involve objects of varying, and often unknown, shapes and sizes.
	Learning to perform manipulation tasks across such a wide range of objects,
	without access to their underlying geometric models, is a challenging problem.
	Recent work has shown how simple keypoint representations can be used to obviate the need of known geometric models \cite{manuelli2019kpam, florence2018dense}.
	These keypoint representations, which are learned purely from visual data, are easy to acquire and provide accurate and robust intra-category generalization capabilities.
	Such keypoint representations have been utilized to formulate optimization problems, whose solutions results in a one-step $SE(3)$ action that is performed by the robot \cite{manuelli2019kpam, florence2018dense}.
	Alternately, they have also been used for state estimation \cite{florence2018dense, ganapathi2020learning}, wherein the keypoints are often directly used as inputs to monolithic neural networks that output the action to be performed at each step.
	
	Instead of using monolithic policies for task learning,
	recent work \cite{sharma2020objaxes}
	has proposed a more modular approach by defining task-axis controllers for each possible subtask.
	These controllers are attached to different objects 
	(or their parts) in the scene, such as the normal of a table or middle of the door handle.
	This object-centric nature of controllers provides important invariances to certain object properties such as a controller that reaches close to an object will be invariant of its position.
	More importantly, these controllers are reusable across multiple different tasks and 
	provide a structured action space for the robot to explore and act.
	This approach results in improved sample complexity and much better generalization for manipulation tasks, 
	and is referred to as \emph{object-centric task-axes controllers}.
	
	\begin{figure}[t]
		\centering
		\includegraphics[width=0.99\linewidth]{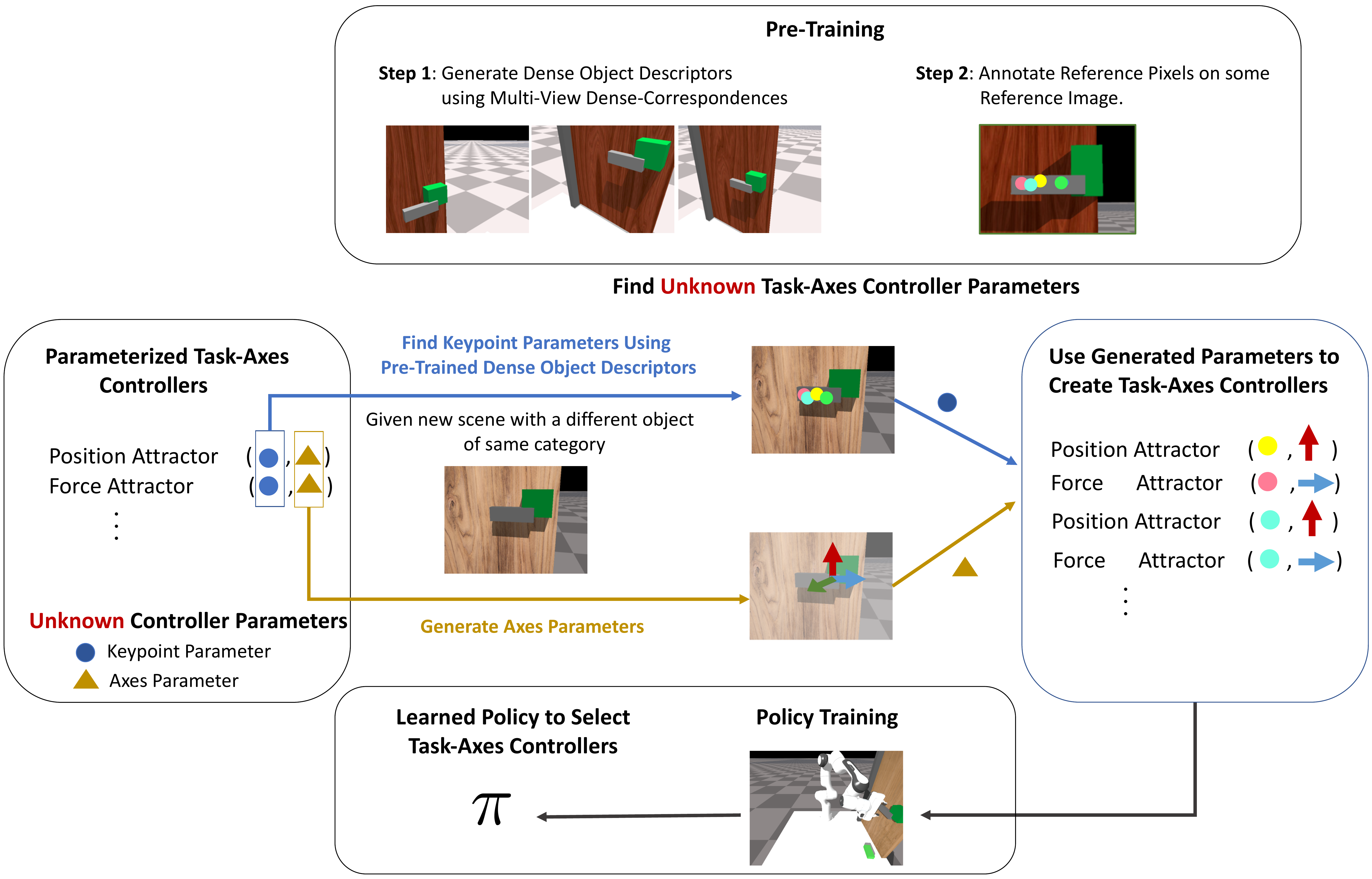}
		\caption{Overview of our proposed approach. We extend task-axes controllers to operate on visual input and use them to present a simple and generalizable approach for learning manipulation tasks.}
		\label{fig:overview}
		\vspace{-5mm}
	\end{figure}

	One limitation of \cite{sharma2020objaxes} is that they do not infer the controller parameters directly from the observed data.
	Instead, they use heuristics to define the set of possible controller parameterizations for each task.
	These controller parameters include both the \emph{position targets}, \emph{i.e.}, 3D positions for relevant objects or 
	other semantically meaningful points on the object such as edges or corners,
	as well as the \emph{relevant axes}, \emph{i.e.}, the axes along which the controller acts.
	
	Our aim in this work is to extend \cite{sharma2020objaxes} to allow it to infer the controller parameters directly from visual input.
	Thus we avoid the use of fixed heuristics to find the position-target parameters for the controllers. 
	This is important since such heuristics are often defined as functions of object parameters, and thus assume direct knowledge of an object's shape, size, and overall geometry, which might not be easily available in the real world.
	Instead, we propose to use keypoint representations based on dense object descriptors to infer these parameters directly from visual data.
	Additionally, instead of using heuristics to provide the axes-parameters we populate them automatically for each of the task-axes controllers.
	This results in a simple approach that allows the robot to learn complex manipulation tasks directly through interaction. 
	
	Our overall contributions include:
	1) We extend object-centric task-axes controllers to infer the controller parameters directly from visual input.
	2) Since learning both task-specific controller parameters and task-specific controller combinations (i.e. task-policy) together is a challenging problem, 
	we show how our proposed approach solves this problem by learning to bootstrap controller parameters using dense correspondence learning.
	3) We empirically validate our approach on multiple manipulation tasks and show its generalization abilities across objects with different shape, size and geometry.
	Video results for all tasks can be accessed \href{https://sites.google.com/view/robotic-manip-task-axes-ctrlrs}{at this link.}


	\section{Related Work}
	\label{sec:rw}
	
	\textbf{Task Frames:} 
	Task frames (or task-spaces) have long been long used by robotics community for robust task execution. 
	Early works of \cite{mason1981compliance, raibert1981hybrid, ballard1984task} formalized and used the notion of task-axes and constraint based task-frames for different manipulation tasks. 
	For robust task execution, roboticists often design specific motions relative to some fixed task-frame or task-axes 
	\cite{muhlig2009automatic, berenson2011task, king2016rearrangement, kober2015learning, ureche2015task, migimatsu2020object, manschitz2020learning}.
	More recent works, have also proposed techniques to learn to select the appropriate task frame for the given task \cite{muhlig2009automatic, kober2015learning, ureche2015task, peternel2017method, conkey2019learning}. 
	These methods use Imitation Learning (IL) combined with manually defined heuristics such as inter-trial variance between human demonstrations to rank proposed task-frames.
	Recent work \cite{sharma2020objaxes} have also proposed using Reinforcement Learning (RL) to choose 
	multiple different controllers both sequentially and in parallel to complete a task. 
	Each controller in \cite{sharma2020objaxes} is defined with respect to some task axes or target keypoint.
	To avoid controllers at each step from interferring with each other null-space projections are employed.
	In this work, we further extend this line of research by learning controller targets and controller axes from
	visual observations.
	We combine task-axes controllers with recent advancements in multi-view correspondence learning to show how 
	the resulting technique can generalize to a large set of scenarios.

	\textbf{Multi-view Correspondence Learning:}
	We use keypoints to define targets for the different task-axes controllers.
	For this we build upon the recent work on using keypoints for manipulation learning \cite{manuelli2019kpam, florence2019self, florence2018dense, Gao_kpam2}. 
	Most closely related to our work is \cite{manuelli2019kpam} which uses supervised learning to detect relevant
	keypoints on objects, which are then used to define a specific optimization problem for each task. 
	Solving this optimization problem leads to a $SE(3)$  transformation which is executed to perform the task. 
	In work concurrent to ours, kPAM \cite{manuelli2019kpam} was extended to kPAM 2.0 \cite{Gao_kpam2}.
	kPAM 2.0 uses oriented keypoints, i.e., keypoints with local orientation information and actions that are defined with respect to these keypoints. 
	This leads to a modular architecture similar to ours.
	However, the control policy in \cite{Gao_kpam2} is manually specified or uses demonstrations while we use more general task-axes controllers and learn their composition using RL.
	Further, as we show once this RL policy is learned it can also be used to refine/learn keypoint representations for a new set of objects thus completing the perception/control loop.
	Alternately, \cite{qin2019keto} learn task-specific keypoints to solve tool manipulation tasks.
	In \cite{qin2019keto} tool manipulation is posed as an optimization
	problem which uses the inferred keypoints. 
	Although \cite{qin2019keto} does not use annotations, they rely on a manually defined heuristic policy to initially train the keypoint generation network.
	By contrast, we use multi-view correspondence learning \cite{schmidt2016self} \cite{florence2018dense} to learn dense object descriptors which are used to infer keypoints.
	Also, instead of a heuristic policy, we rely on a single human annotation, which when combined with dense object descriptors works well in practice.
	More importantly, unlike \cite{manuelli2019kpam, qin2019keto} which solve optimization problems, we instead solve a temporally extended RL problem, 
	where at each step which task-axes controller to execute needs to be learned by interaction.
	
	\textbf{Learning with Parameterized Actions:}
	Learning to select the task-axes controller to execute at each step (task-policy), as well as the target and axes parameter for this controller is closely related to learning with parameterized actions.
	Parameterized action spaces consider the problem where each action of an MDP is parameterized by a low dimensional input.  
	In \cite{masson2016reinforcement}, parameterized action-MDPs were referred to as PAMDPs.
	PAMDPs are challenging to solve since they require a
	bilevel optimization algorithm, 
	wherein the outer loop searches over the continuous action
	parameters, while the inner loop optimizes for the discrete action selection to complete the task. 
	Since the inner loop requires solving an RL problem, this bilevel optimization is challenging to perform for high-dimensional spaces.
	Alternative works \cite{hausknecht2015deep} have been proposed that avoid this bilevel optimization problem by exploring both, the parameter-space and the action-selection space together.
	However, since these works explore using uniform distributions over the action parameters, they are unsuitable when these parameters need to be inferred from high dimensional spaces, \emph{e.g.}, selecting a pixel (keypoint) from an image.
	In current work, we avoid the computational complexity of this bilevel optimization by bootstrapping the controller parameters using multi-view correspondence learning.

	\section{Overview}
	
	Manipulation tasks often involve different objects and manipulating them through contact to achieve desirable effects.
	The use of object-centric task-axes controllers provides a structured action space for the robot to both explore and act in.
	However, as discussed previously, instead of using heuristics to specify the parameters for these task-axes controllers, we would like to infer them directly using perceptual input.
	In the following sections we first discuss the task-axes controllers formulation based on previous work, along with the different types of controllers that are used and their associated parameters.
	Following this, we discuss our approach to learn these parameters directly for each task.
	
	\section{Object-Centric Task Axes Controllers}
	
	Prior work \cite{sharma2020objaxes} defines multiple different task-axes controllers, each associated with some underlying object in the scene. Each controller is also associated with some predefined task-axes.
	These task-axes are centered at some task-specific target positions, 
	such as object centers or other semantically useful points on the object \emph{e.g.} edges or corners.
	Further, these task-axes can be oriented in different ways \emph{e.g.} along the surface normal, or along some world axes. 
	In the following sub-sections, we briefly discuss some of the relevant controllers used in our current work and subsequently detail their associated parameters.
	
	\subsection{Controller Types}
	\label{subsec:controller_types}
	
	We use three different types of controllers --- position, force, and rotational alignment controllers. 
	We use position controllers as attractors that move the end-effector (EE) close to some target position on an object of interest. 
	We also use another set of position controllers, which are implemented as curl attractors, that allow us to rotate around objects of interest to align with them.
	In addition to position controllers, we also use force controllers which apply a fixed force magnitude along a particular target axes, \emph{e.g.},
	pushing an object down requires a force along its vertical direction.
	Rotation controllers are used to align end-effector axes with some given object axes. 
	The position and force controllers generate delta translation targets for the end-effector, and the rotation controllers generate delta rotation targets.
	
	We now discuss the exact set of parameters used for each controller defined above. 
	We use $x_c \in \mathbb{R}^3, R_c \in SO(3),$ and $f_c \in\mathbb{R}^3$ to denote the \emph{current} end-effector translation, rotation, and forces expressed in the robot's base frame.

	\textbf{Position and Force Controllers:}
	The position controller is parameterized by a target position $x_d$ and an axis $u$ along which the controller will move the robot's end-effector toward the target.
	$u$ can either be a fixed direction, like the normal direction of a surface, or it can be time varying while following $x_d$, \emph{i.e.}, 
	$u = \frac{x_d - x_c}{\|x_d - x_c\|_2}$.
	The translation error for the above controller can be computed as, \emph{i.e.}, $\delta_x(x_d, u, x_c) = \mathcal{P}(u) (x_d - x_c)$, where $\mathcal{P}(u) = u u^\top$ is the projection matrix for the axis $u$.
	The force controller is also parameterized by a force target $f_d$ and a direction $u$, which can also be fixed or time-varying (as defined above).
	
	\textbf{Rotation Controller:} is parameterized by unit vector $u$, which selects an EE-axis using $R_cu$ and aligns it with a target axis $r_d$. Together, $(u, r_d)$ parameterize the rotation controllers.
	The rotation controller produces the following delta rotation target (using the angle-axis representation), $\delta_R(r_d, u, R_c) = \cos^{-1}( (R_c u)^\top r_d) ((R_c u) \times r_d)$.
	
	We refer the reader to \cite{sharma2020objaxes} for more details on these controllers as well as their visualization and implementation.
	
	\textbf{Controller Parameters:}
	From the above discussion we get the following parameters for each controller, $(x_d, u)$ for position controllers, $(f_d, u)$ for force controllers 
	and $(r_d, u)$ for rotation controllers.
	Finally, since each controller is either implemented as a PD, PI or PID controller, they also require suitable gain parameters $K_p$, $K_d$, $K_i$.
	In this work we focus on learning parameters that can be directly inferred from visual input, \emph{i.e.}, the 3D position targets $(x_d)$, rotation target $r_d$ and target-axes $(u)$.
	While we fix the values for other parameters.
	In the following discussion, we use the term learned controller parameters to refer to this initial set of parameters only.


	\section{Learning Controller Parameters}
	
	In prior work, desired position targets $x_d$ are defined using 
	task-specific heuristics such as middle of the door handle or the middle of the wall as target positions. 
	Task-specific knowledge is also used for task-axes parameter $u$ and rotation targets $r_d$.
	For instance, for the door opening task we only add rotation controllers that align the downward end-effector axes with the door normal.
	By contrast, we want to infer these parameters from visual input and learn the appropriate parameters based on task interaction instead of heuristics or apriori information.
	
	As noted previously, to learn both --- relevant $x_d$, $u$, $r_d$ for each task-axes controller as well as the composition of these task-axes controllers for task completion, is a challenging problem.
	This is because it requires solving a bilevel optimization problem with an inner loop that involves solving a challenging RL problem \cite{masson2016reinforcement}.
	This challenge is further magnified when the controller parameters have to be inferred from high dimensional input such as images. 
	For instance, parameters such as 3D controller position targets can lie anywhere on the image space.
	
	To overcome these challenges we propose to bootstrap learning the controller position parameters $x_d$ through multi-view correspondence learning and a few human annotations.
	We simultaneously infer a set of candidate axes parameters for both $u$ and $r_d$ from the given objects in the scene.
	Using these bootstrapped parameters we learn a task-axes controller composition policy $\pi$ using RL.
	With this learned $\pi$ we can subsequently learn $x_d$ parameters for novel objects for which no human annotations are available.
	In the following sub-sections we explain the proposed approach in more detail.
	
	
	\begin{figure}[t]
		\centering
		\includegraphics[width=0.50\textwidth]{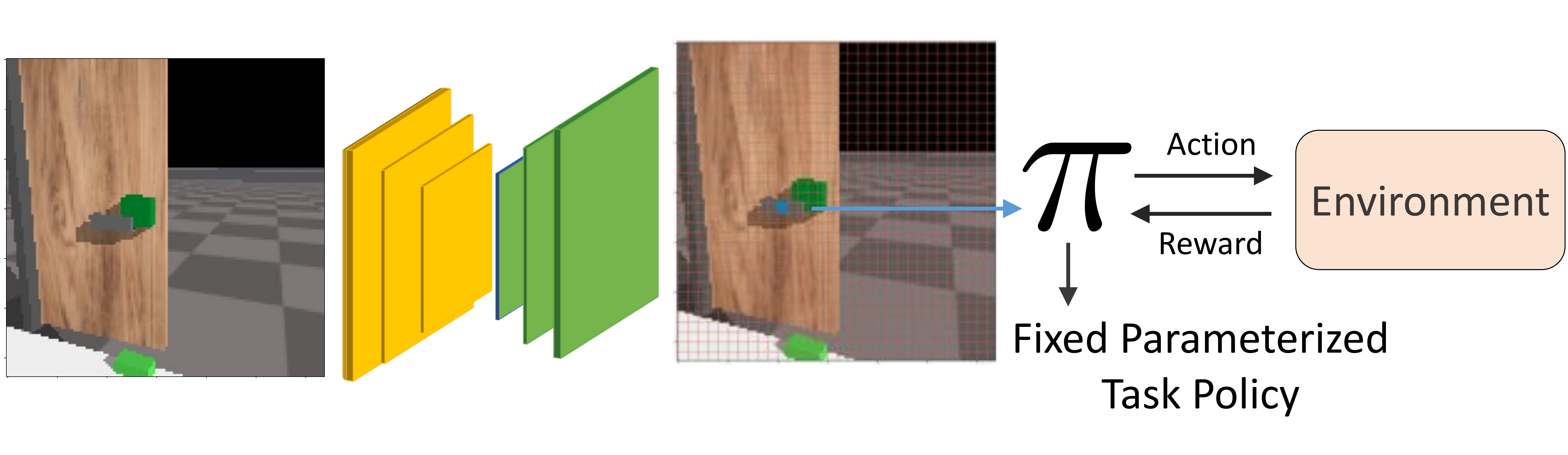}
		\caption{The model used to learn keypoint parameters using
			the bootstrapped task policy. Detailed model architecture is in Appendix~\ref{app:keypoint-model}.}
		\label{fig:keypoint-model}
		\vspace{-4mm}
	\end{figure}

	\subsection{Learning to Bootstrap Controller Parameters}
	\label{subsec:learning-bootstrap}
	
	We now discuss our approach to bootstrapping the position target and target-axes parameters. First, we look at the case of 3D position parameters $x_d$, and subsequently we look at the axes parameters $u$ and $r_d$.
	
	\subsubsection{Keypoint Parameters}
	
	We refer to the 3D position target parameters $(x_d)$ of each controller as keypoint parameters.
	These keypoint parameters are associated with semantically meaningful parts of underlying objects such as middle of the door handle, center of the button, edge of a block.
	To bootstrap learning the keypoint parameters, we propose to use 
	multi-view dense correspondence learning.
	Specifically, we use DenseObjectNets \cite{florence2018dense, florence2019self} which learn dense object descriptors for each pixel from multi-view data in a completely self-supervised manner. 
	Not only does this setup avoid the need of any expensive manual data-labeling procedure,  
	prior works have shown that the learned object descriptors are quite robust to the presence of mild occlusions and importantly, lead to category-level generalizations \cite{florence2018dense}.
	As we show empirically in Section~\ref{sec:results}, such category level generalization allows us to infer consistent controller targets irrespective of the object's size and position as well as variation in its shape and geometry.
	
	To train dense object descriptors, 
	we use a small set of objects ($\approx 10$) relevant to each task family and learn dense descriptors on these objects. 
	All of these objects belong to the same category, \emph{e.g.}, for the door opening task we learn descriptors across doors with varying sizes of door handles as well as their locations on the door frame. 
	Thus, the learned dense object descriptors should generalize to other novel objects that belong to the same category, \emph{e.g.}, door handles with more complex shapes. Figure~\ref{fig:results_door_handle_correspondences} visualizes some learned object descriptors.
	We refer to this dense-object network model as our \emph{bootstrapped} keypoint model $\phi$.
	
	Given $\phi$, we infer the keypoint parameters for each controller by using a \emph{reference} image $(I_r)$ from the dataset collected for training dense descriptors. 
	This reference image is used as representative for the object category being manipulated. 
	To extract keypoints we manually label a set of reference pixels on $I_r$, denoted as $P\coloneqq\{p^1_r, p^2_r, \cdots\}$.
	These reference pixels $p^i_r$ encode semantic information about the scene which is relevant for the task.
	For instance, for the door open task we label pixels near the edge and middle of the door handle since these keypoints afford grasping and rotating the door handle compared to pixels closer to the handle's rotation joint.
	Similarly for the block tumble task we label pixels near the edge instead of the middle of the block.
	Figure~\ref{fig:results_keypoint_params} shows example keypoints for the tasks considered in our current work.
	Assuming known camera parameters we get 3D position targets $\{x^1_d, x^2_d, \cdots \}$ from $\{p^1_r, p^2_r, \cdots\}$ directly.
	To get position targets for a new image, which contains a novel object of the same category as $I_r$, we use pixels that are closest to the reference pixels in the descriptor space \emph{i.e.}, $p^i = \argmin_{p} \| \phi(p^i_r) - \phi(p) \|_2$, where $p$ is any pixel on the image, and $p^i$ is used to get the \emph{i'th} controller target $x^i_d$.

	\begin{figure}[t]
		\centering
		\includegraphics[width=0.48\textwidth]{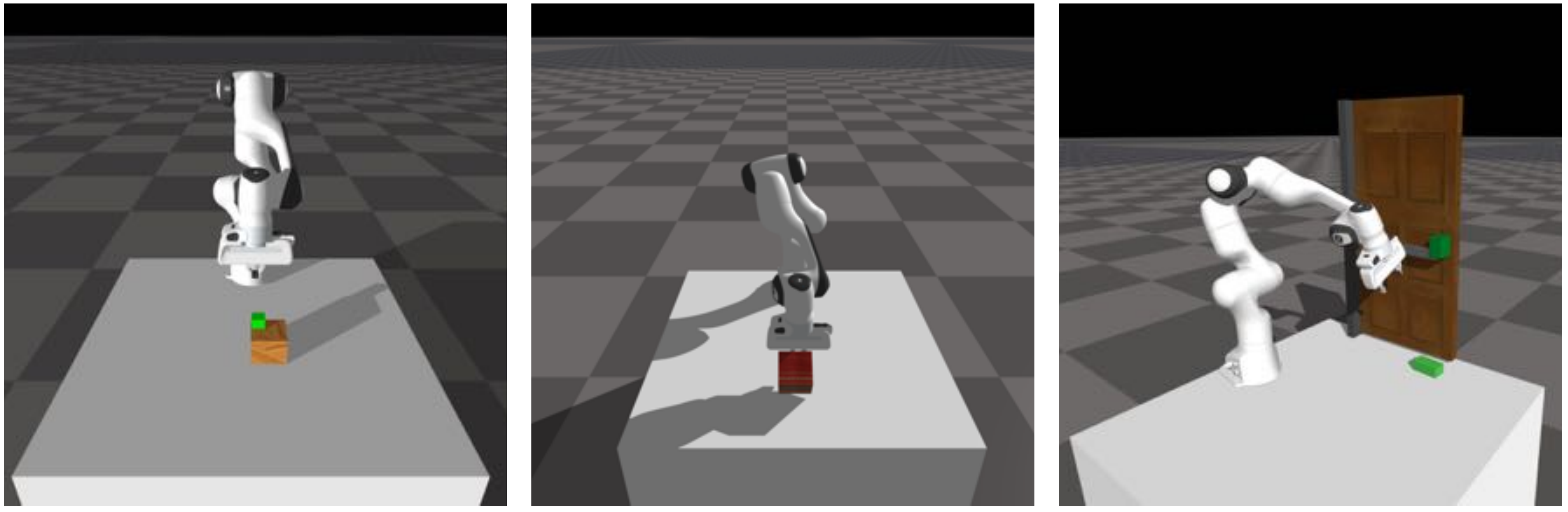}
		\caption{The three different tasks used to evaluate our proposed approach. 
			From \emph{left} to \emph{right}: Button Press, Block Tumble, and Door Opening.}
		\label{fig:tasks}
		\vspace{-3mm}
	\end{figure}
	
	\subsubsection{Axes Parameters}
	
	Both rotation target $r_d$ and target-axes $u$ parameters require valid 3D axes.
	To find these relevant axes for each controller, we initially extract a relevant set of 3D axes for the given scene.
	We refer to this as the \emph{candidate axes} set, $A \coloneqq \{a^1, a^2, \cdots\}$.
	There exist multiple approaches to extract these candidate axes.
	For instance, 
	possible candidate axes include both the global (world) axes as well as object axes.
	Additionally, we can also extract candidate axes by using the local geometry of the object around each inferred keypoint \emph{e.g.}, using the axes normal to surface or along the surface.
	In this work, we directly use the object axes and the global axes as our candidate axes.
	
	Given $A$ the set of candidate axis, we can find $u$ for each controller by finding the most relevant axes from $A$.
	However, this requires task-specific knowledge \emph{e.g.}, in the form of user defined priors as used in previous work.
	Instead, we avoid this by associating each position target with every axis $a^j \in A$.
	Additionally, each axes in $A$ can be used to create a unique rotation target $r_d$. 
	One drawback of this approach is that it results in a combinatorial number of controllers, since we associate every position target with each axes in $A$.
	This can result in a large action space for $\pi$ to explore.
	In Section~\ref{sec:results}, 
	we empirically validate how this choice affects the sample complexity of using task-axes controllers for different manipulation tasks.
	Figure~\ref{fig:overview} visualizes the pipeline of our overall approach. 
	
	\subsection{Learning Controller Parameters via Learned Task-Policy}
	\label{subsec:learn-controller-params-learned-task-policy}
	
	Using the above approach we can bootstrap controller parameters and learn task-specific manipulation policy $\pi$.
	This policy learns to compose different task-axes controllers to achieve the 
	overall manipulation task.
	As we show empirically, this combination of bootstrapped controller parameters and learned task-policy can now be used on novel objects with varying geometry, shape, size etc. 
	This impressive level of generalization is a result of both, the category level generalization provided by dense-keypoint parameters and the object-centric nature of task-axes controllers.
	
	\begin{figure}[t]
		\centering
		\includegraphics[width=0.95\linewidth]{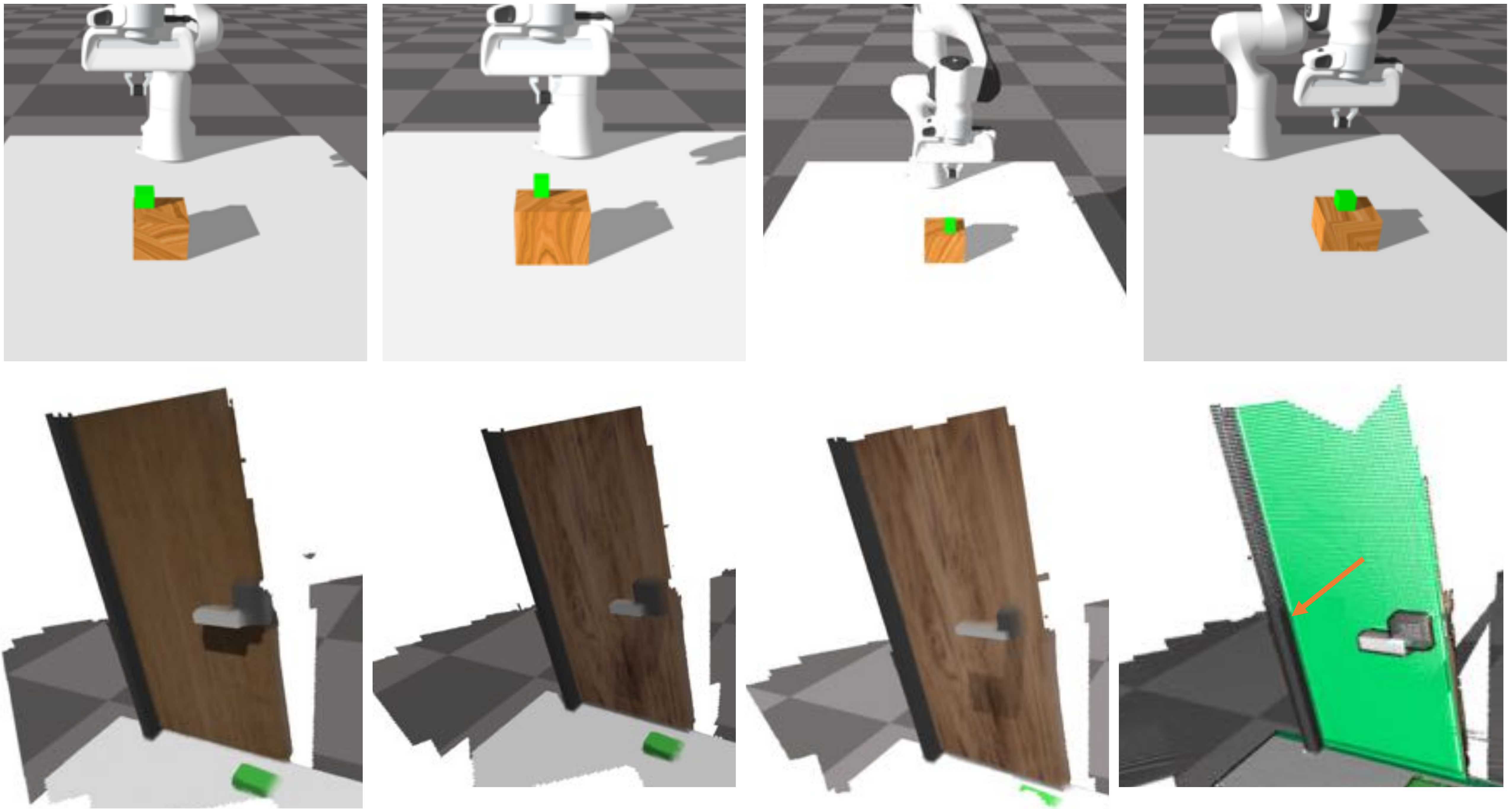}
		\caption{Example task variations used for training the Button Press and Door Open tasks.}
		\label{fig:button_task_variation}
		\vspace{-3mm}
	\end{figure}
	
	However, in some scenarios we may want to use the learned task policy on objects 
	that belong to a different category as compared to objects that were used to learn the bootstrapped keypoint parameters.
	For such cases, the keypoints $p^i$ inferred by the bootstrapped dense-object network based keypoint model $(\phi)$ can be unsuitable.
	In such cases, instead of re-training the dense-object network on this new set of objects and relying on human annotations to get corresponding reference pixels $P$,
	we can utilize the learned task policy $\pi$ and learn the mapping to keypoint parameters directly from image space.
	This is possible because of two reasons. 
	First, since $\pi$ is implicitly parameterized by the controller target parameters $x_d$, 
	we can learn $x_d$'s value for the new set of objects based on the feedback from rolling out $\pi$.
	Second, this assumes that task policy $\pi$ is approximately similar across both sets of objects, 
	\emph{i.e.}, the set of objects used during bootstrap learning and the new set of objects. 
	This assumption holds for the tasks we consider, 
	\emph{e.g.}, to learn a door opening policy, we only need to reach close to the handle, grasp it from a location which affords grasping, rotate the handle and pull it. 
	
	
	To learn keypoint parameters for new set of objects,
	we learn a new neural network $\psi_\theta$, with weights $\theta$, 
	by rolling out $\pi$ and learning to optimize the underlying task reward 
	$J(\theta) = \mathbb{E}_{\tau\sim\pi(\cdot\vert s, \psi_\theta(I_s))}\left[ r(\tau) \right] $,
	where $\psi_\theta(I_s)$ denote the keypoint parameters inferred from image $I_s$ at state $s$, and $\pi$ represents the learned task policy. 
	We do not update $\pi$ during this keypoint learning stage.
	$J(\theta)$ can be optimized using any policy gradient method \emph{e.g.} Reinforce\cite{williams1992simple}, 
	$\nabla J(\theta) = \mathbb{E}_{\tau}\left[r(\tau)\nabla\log \psi_\theta(s) \right]$.
	We represent $\psi_\theta$ using a fully-convolutional network, with the last layer of the network containing as many channels as the number of keypoint parameters to infer.
	We use a softmax over each last channel and sample the appropriate pixel to get $p^i$ which is then used to infer the controller target $x^i_d$.
	Figure~\ref{fig:keypoint-model} provides an overview of our approach, 
	while Appendix~\ref{app:keypoint-model} provides more details about the network architecture and its training.

	\section{Experimental Setup}
	
	With our experiments we aim to evaluate:
	1) How well does our proposed approach using bootstrapped controller parameters perform?
	Since a combinatorial mix of keypoint and axes parameters results in a large action space,  
	we investigate how this affects overall task performance and sample complexity of our approach (Section~\ref{subsec:learning-bootstrap} Training).
	2) How well does the combination of dense-object net based keypoint model and 
	object-centric task-axes controller generalize to new objects of varying shapes, sizes and geometry? (Section~\ref{subsec:learning-bootstrap} Generalization),
	3) How well can a learned task policy be utilized to learn keypoint controller parameters for new set of objects (Section~\ref{subsec:learn-controller-params-learned-task-policy})?


	\begin{figure*}[t]
		\centering
		\includegraphics[width=0.9\textwidth]{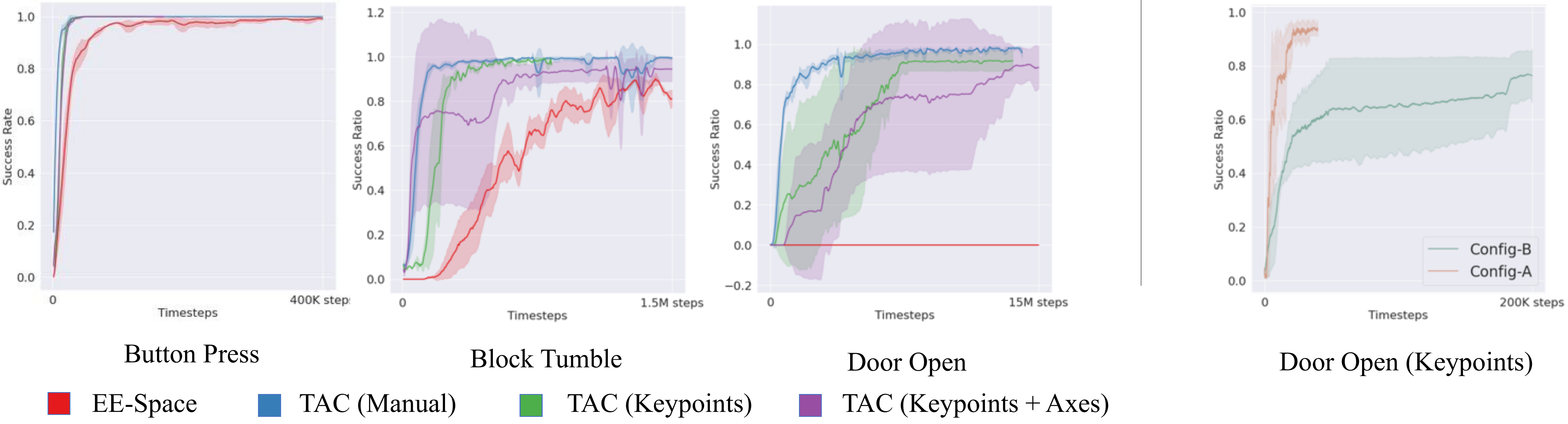}
		\caption{\emph{Left:} Task Success Rate for all 3 environments with bootstrapped controller parameters. 
			\emph{Right:} Task success rate on two different configs (door-handle types) for learning keypoints using a learned policy for the Door-Open Task. 
			Figure~\ref{fig:results_refine_keypoints_1} visualizes both configs.
			The dark line shows mean sucess-ratio, shaded region plots std across 5 seeds.
		}
		\label{fig:task_success}
		\vspace{-3mm}
	\end{figure*}
	
	\subsection{Tasks}
	
	To evaluate our proposed approach we use 3 different manipulation tasks (Figure~\ref{fig:tasks}) of increasing complexity -- Button Press, Block Tumble, and Door Opening.
	
	\textbf{Button Press:}
	In this task, a 7-DoF Franka Panda arm is used to push down a button which is positioned on top of a box placed infront of the robot (Figure~\ref{fig:tasks} left).
	Instead of using only one type of button object, we verify our approach on multiple objects with different shapes and sizes.
	These variations include the button position on top of the box object, sizes of both the button as well as its underlying box object. Figure~\ref{fig:button_task_variation} (top) visualizes some of these variations.
	
	\textbf{Block Tumble:}
	In this task, a 7-DoF Franka Panda arm is required to tumble a block along a particular axis 
	(Figure~\ref{fig:tasks} middle).
	This task is particularly interesting since there exist multiple different ways to accomplish it. 
	For instance, the robot can tumble the block by applying a downward force anywhere along its edge.
	To test generalization for this task, we vary the size of the block between 0.07m to 0.16m.
	
	\textbf{Door Opening:}
	We also verify our approach on the door-opening task.
	In this task, the Franka robot needs to open a door by first turning its door handle and then pulling the door beyond an opening threshold.
	In contrast to \cite{sharma2020objaxes}, 
	which tests generalization by only varying the position of the door handle on the door frame, 
	we vary both the size of the door handle as well as the location of the door handle on the door. 
	Additionally, we also change the door shape and evaluate the proposed approach for both cuboidal, cylindrical and more complex door handle shapes.
	See Figure~\ref{fig:button_task_variation},\ref{fig:results_door_gen_1} or attached video (project page) for reference.

	\subsection{Compared Approaches}
	We compare our proposed approach against multiple different methods. In the below sections we refer to task-axes controllers using the shorthand TAC.
	\begin{enumerate}
		\item \textbf{EE-Space:} 
		We verify the utility of the structured action space provided by object-centric task-axes controllers by comparing against an approach that directly controls the robot via end-effector delta targets.
		\item \textbf{TAC (Manual)} We also evaluate our approach against manually specified controller parameters. 
		We note that we do not aim to use the \emph{smallest} possible number of controllers and their parameters. 
		Instead, we specify the controllers and their parameters only to provide some useful priors for overall task learning. 
		\item \textbf{TAC (Keypoints):} We evaluate one version of our proposed approach in which we only infer the keypoint parameters \emph{i.e.} the target positions for each position or force controller.
		We reuse the axes specified for TAC (Manual) with the inferred keypoints.
		\item \textbf{TAC (Keypoints+Axes):} 
		We evaluate our proposed approach wherein both the keypoints as well as the axes parameters are inferred for each scene.
	\end{enumerate}

	\begin{table}[]
		\centering
		\resizebox{\linewidth}{!}{%
			\begin{tabular}{@{}llll@{}}
				\toprule
				Env & EE-Space & TAC (Manual)* & TAC (Keypoints+Axes) \\ \midrule
				Button Press & 0.98 (0.01) & 1.0 (0.0) & 1.0 (0.0) \\
				Block Tumble & 0.487 (0.26) & 0.96 (0.03) & 0.932 (0.04)  \\
				Door Opening & 0.0 (0.0) & 0.97 (0.01) & 0.94 (0.05) \\ \bottomrule
			\end{tabular}%
		}
		\caption{Mean (std) for each method on all three different manipulation tasks. Each approach was verified on 15 different environment variations.
		}
		\label{tab:results_quant_generalization}
		\vspace{-4mm}
	\end{table}

	Table~\ref{tab:controller_stats} shows the number of keypoints and axes parameters inferred for each of the tasks. 
	We note that there exist controllers which do not have any axes associated with them such as the error axis controllers.
	
	\textbf{Metrics:}
	We show qualitative and quantitative results for two scenarios.
	First, we show results for learning task policy using bootstrapped controller parameters.
	Second, we use this learned task policy to show results for learning controller parameters using direct interactions.
	For both scenarios we compare approaches using the success ratio metric. 

	\begin{figure*}[t]
		\centering
		\includegraphics[width=0.86\textwidth]{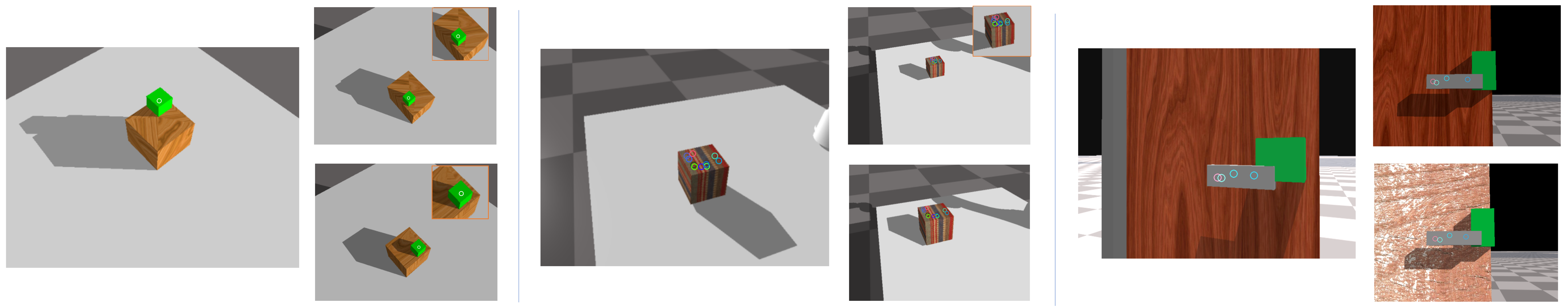}
		\caption{
			Visualization for \emph{reference} images and pixels (left image in each column) and corresponding pixels
			predicted using learned descriptors. For door open, one reference pixel is closer to door joint to show that our approach learns to not use it.}
		\label{fig:results_keypoint_params}
	\end{figure*}

	\textbf{RL Training:} 
	For training the task policy we use Proximal Policy Optimization (PPO) \cite{schulman2017proximal} based on stable-baselines \cite{stablebaselines}. 
	While for learning controller parameters we found a simple Reinforce \cite{williams1992simple} based approach to be sufficient.
	All results are run and reported for 5 different seeds.
	Hyperparameters for each setting are reported in the Appendix~\ref{app:train_details}.

	
	\section{Results}
	\label{sec:results}

	\subsection{Learning Task Policy}
	
	\subsubsection{Training Results}
	Figure~\ref{fig:task_success} plots success ratio for all approaches across each task.
	As seen above, we observe that for the simplest task \emph{i.e.} button press, all methods are able to learn the task quickly. 
	Also, since the underlying task is not complex, each method has little variance across multiple seeds.
	For the Block Tumble task (Figure~\ref{fig:task_success} middle), although all methods perform well on the training task, 
	methods that use task-axes controllers are much more sample efficient.
	This is true even when we use a much larger set of controllers \emph{i.e.} the TAC (Keypoints + Axes) approach.
	This is because most of the keypoints used in the task 
	(Figure~\ref{fig:results_keypoint_params}) can be used to accomplish the task. 
	Also, since there is only one axes along which the block needs to be flipped, 
	the robot is quickly able to find this relevant axes using the provided dense rewards.
	Thus, even with a much larger action space our approach of inferring the keypoints and axes 
	performs similarly, when compared to their manual specification.

	Figure~\ref{fig:task_success} (right) plots the success ratios for the door opening task. 
	From the above figure, we observe that the EE-space is unable to solve the overall task. 
	Similar results were also observed in \cite{sharma2020objaxes}.
	The main reason for this failure is the overall task complexity, especially since task completion requires many different subtasks (reaching, grasping, turning the handle and pulling it back) to be performed in sequence.
	On the other hand, we observe that all methods that utilize task-axes controllers are able to learn to perform the task.
	However, in contrast to the previous tasks, our proposed approach, TAC (Keypoints+Axes), does require more samples compared to when we manually provide these parameters, TAC (Manual).
	We also observe that only inferring the keypoints and not the axes (TAC - Keypoints) is still quite sample efficient.
	This indicates that the agent in TAC Keypoints+Axes does spend some initial time exploring different axes which can be used to accomplish the task.
	This is possible because there exist multiple ways to grasp the handle. 
	For instance, it is possible to grasp the handle both along the vertical as well as the horizontal axes. However, using the vertical axes is not robust, since it can easily collide with the door frame.
	Thus, as a large number of actions are not particularly useful for the task, the agent will have to interact and learn the most suitable and robust ways to achieve it. 
	
	\subsubsection{Generalization Results}
\begin{figure}[t]
		\centering
		\includegraphics[width=0.96\linewidth]{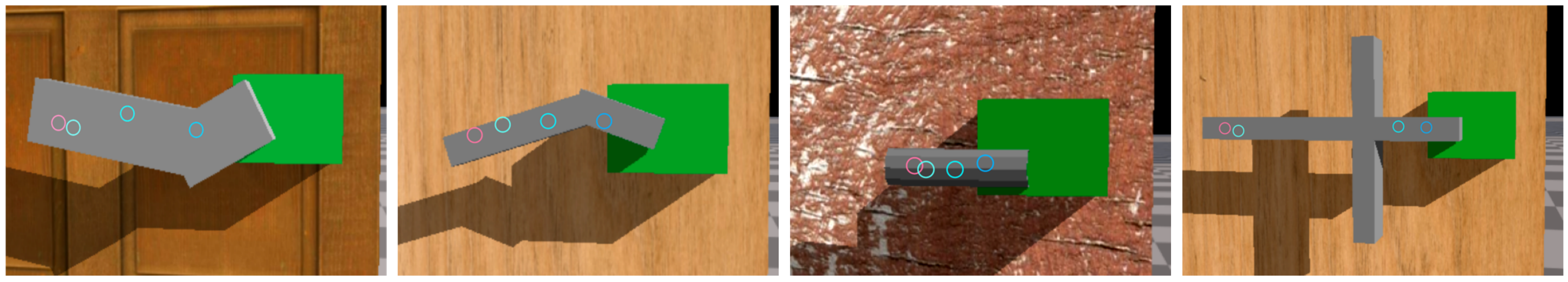}
		\caption{\emph{Qualitative Results} we show that our learned control policy although not trained on any of the above models does successfully transfer to them (see results in project-page).}
		\label{fig:results_door_gen_1}
		\vspace{-3mm}
	\end{figure}

	Table~\ref{tab:results_quant_generalization} shows the generalization performance for three different methods.
	This generalization performance was recorded on $15$ different environment settings with varying object sizes and shapes.
	We note that for TAC (Manual) we only used primitive shapes (cuboids and cylinders) since we need to manually provide keypoint parameters. 
	By contrast, for TAC (Keypoints+Axes), we directly use the
	visual input to infer the relevant keypoint parameters.
	See Figure~\ref{fig:results_door_gen_1} for test configurations used for TAC (Keypoints+Axes), for door open task.
	As seen in Table~\ref{tab:results_quant_generalization},
	both methods that use task-axes controllers are able to generalize quite well across all of the tasks.
	On the other hand, the EE action-space provides good generalization capabilities only for the simplest task (Button Press). 
	While even for the moderately complex task of Block Tumble its performance reduces significantly. 
	One reason for this drop in performance is the lack of any inductive bias in the EE-space, 
	and since we train on a small set of objects only, the EE-space policy fails to generalize to larger changes in object variations.
	Figure~\ref{fig:results_door_gen_1} shows some objects with complex underlying geometry that were never used either for dense descriptor or policy training.
	While TAC (Manual) cannot be applied to such objects, we show in the attached video 
	that our method is successfully able to zero-shot generalize to such large variations as well.

	\begin{figure}[t]
		\centering
		\includegraphics[width=1.0\linewidth]{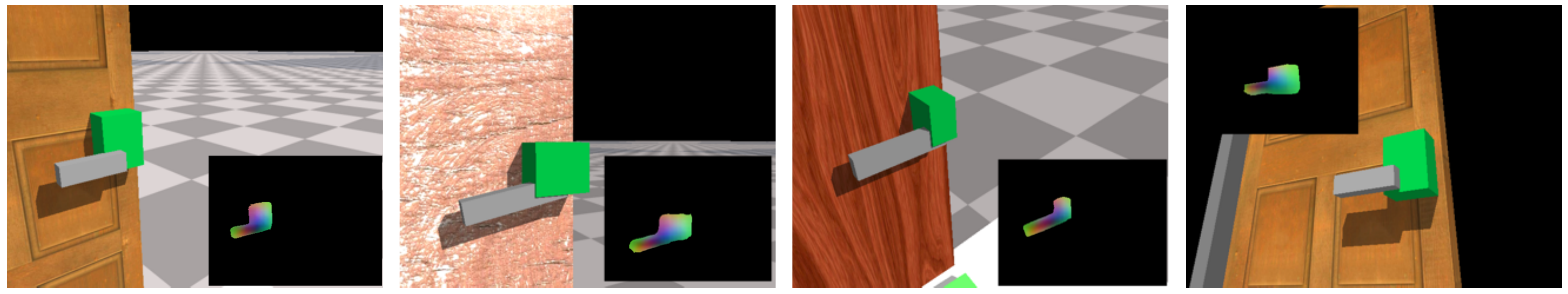}
		\caption{Dense object descriptor results for Door Open Task.}
		\label{fig:results_door_handle_correspondences}
	\end{figure}
	
	\begin{figure}[t]
		\centering
		\includegraphics[width=0.96\linewidth]{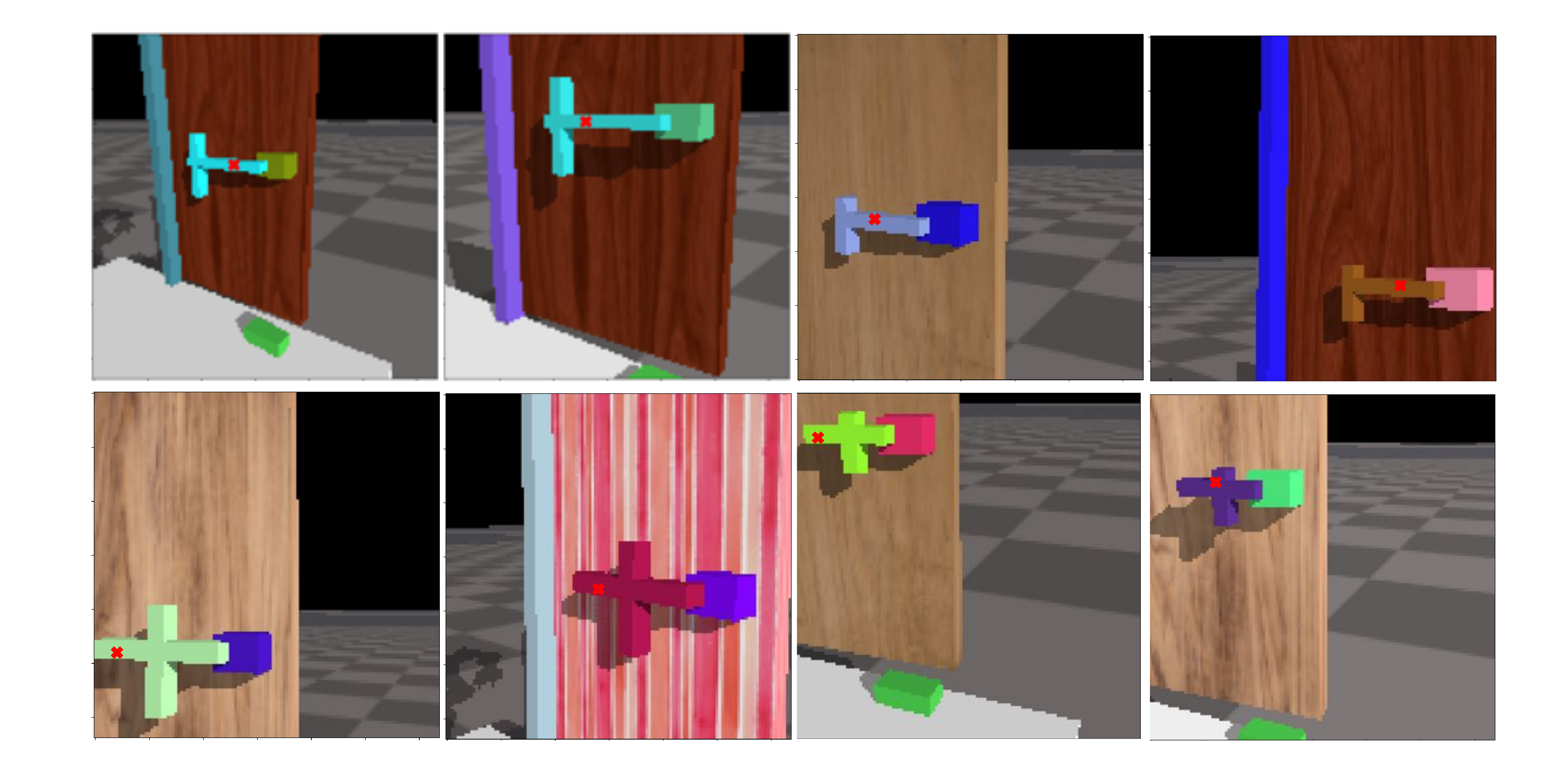}
		\caption{Keypoint Parameters learned using a learned task policy (Section~\ref{subsec:learn-controller-params-learned-task-policy}). Top Row is Config-A, bottom row is Config-B. }
		\label{fig:results_refine_keypoints_1}
		\vspace{-4mm}
	\end{figure}
	
	\subsubsection{Qualitative Results:}
	In addition to the above quantitative results, we also show some qualitative results for the learned descriptors and the inferred keypoints.
	Figure~\ref{fig:results_keypoint_params} plots the keypoints used for each of the tasks. 
	The left image in each column is the \emph{reference} image with the annotated \emph{reference} keypoints.
	While the right column shows scenes with two \emph{different} objects used in the test set to evaluate the learned policies. 
	We visualize objects at same scale to show their original sizes.
	Additionally, in
	Figure~\ref{fig:results_door_handle_correspondences} we plot the learned descriptors for the door opening task. 
	As seen above, the  learned descriptors are able to approximately cluster semantically meaningful regions together. For instance, the part of door handle close to its rotation joint, the middle and end of the door handles, as well as the right end of the hinge are all well estimated.
	While in Figure~\ref{fig:results_door_gen_1} we see that the reference pixels are also able to generalize to objects with very different shapes and geometry.
	We also show qualitative results for these samples in our video results. 
	This is not surprising since as long as the above keypoints afford grasping and rotating the door handle, the underlying task-axes controllers should be able to generalize.
	See video results for all tasks and supplementary material at \url{https://sites.google.com/view/robotic-manip-task-axes-ctrlrs}.
	
	\subsection{Learning Controller Parameters with Learned Task Policy}
	\label{subsec:learn-controller-parmas-with-task-policy}
	
	Figure~\ref{fig:task_success} (right) shows results for learning keypoints using the learned task policy. 
	Since the door opening can be performed with only 1 keypoint target, we only
	use one keypoint for this task policy.
	This task policy is trained on simple door handles only, as visualized in Figure~\ref{fig:results_door_handle_correspondences}.
	While the keypoint learning policy is trained on 
	two completely different door-handle configurations, visualized in Figure~\ref{fig:results_refine_keypoints_1}.
	Additionally, as seen in Figure~\ref{fig:results_refine_keypoints_1},
	we note that for both configurations we ensure that the door handle can lie anywhere on the image space and thus may not necessarily be near the center of the image or always orthogonal to the camera. 
	This makes the keypoint learning problem much more challenging.
	
	Although both configurations (config-A and config-B) are visually quite similar, they result in very different sample complexities when learning to select the appropriate keypoint.
	This is because for config-A there exists a much larger region of the door handle that can be selected to perfectly execute the learned parameterized task-policy.
	This can be seen in Figure~\ref{fig:results_refine_keypoints_1} (top-row) where different keypoints along the door handle have been selected to successfully open the door.
	While for config-B this valid region is quite small (to the left of the vertical handle bar). 
	Additionally,  small inconsistencies in keypoint predictions, 
	such as visualized in Figure~\ref{fig:results_refine_keypoints_1} (bottom-row right) fail to open the door.
	This makes the keypoint learning problem much more challenging, which consequently results in a much larger sample complexity.
	
	Additionally, we provide more results in Appendix~\ref{app:results}.
	Specifically, we show that directly using dense correspondence learning and reference pixels does not perform well on config-A door handles visualized above.
	We further report both qualitative and quantitative results for learning keypoint controller parameters for the block tumble task.

	\section{Conclusion}
	
	Our work in this paper leads to a modular architecture which separates perceptual learning from the control policy, \emph{i.e.}, although the control policy acts on perceptual input, both models are trained separately.
	This stream of work is similar to the recently proposed \cite{manuelli2019kpam, Gao_kpam2}, wherein the predicted keypoints are used to solve an optimization problem whose output is used to perform the task.
	Despite differences among approaches, they all lead to an improved interpretability of the learned models. 
	This is a consequence of semantic keypoints in \cite{manuelli2019kpam}, while for our work this is a consequence of both semantic keypoints as well as task-axes controllers which operate on semantic inputs.
	In addition to interpretability, 
	a modular architecture also allows us to reuse the control policy while retraining the perceptual network on a completely different set of objects (\emph{e.g.} different geometries) and still learn to perform the task as long as the semantic keypoints for both object sets provide the required affordances such as grasping, turning, pushing.
	This is in contrast to end-to-end DeepRL approaches where the perceptual network is closely tied with the control policy and it is not possible to update one without updating the other.

	\bibliographystyle{IEEEtran} 
	\input{references.bbl}

\clearpage
\input{appendix.tex}

\end{document}

%% file: references.bbl

%% file: appendix.tex
\appendices

\section{Experiment Task Details}
In the following sub-sections we give details around each of the task used in the paper.

\subsection{Button Press}

\textbf{Observation Space:}
\begin{enumerate}
    \item 7-dimensional robot arm joint angles.
    \item The gripper width. This is assumed to be closed for this task.
    \item 6D pose of the end-effector.
    \item End-effector contact forces.
    \item 3D block position, for the block under the button.
    \item 3D button position. 
    \item The prismatic joint between the block and the button.
\end{enumerate}

\textbf{Reward Function:}
The reward is a function of the distance to the button, the pressing of the button, a time penalty ($0.1$ at each step) to perform the task as soon as possible, and an overall task bonus to complete the task.
Formally, 
let $d_t$ be the euclidean-distance between the end-effector and the button at time step $t$.
The distance reward then is based on distance improvement to the target \emph{i.e.} $d_t-d_{t-1}$.
Let $j_t$ be the value for the prismatic joint between the block and the button.
Let $succ$ define task success, which is defined by the following indicator function $succ = 1\left\{j_t > 0.1\right\}$.
The overall reward then is
\begin{align}
    r_t &= 10\times\left(d_t-d_{t-1}\right) + 10\times\left(j_t-j_{t-1}\right) \\ \nonumber
    & \qquad + 100\times succ - 0.1 \label{eq:reward_button_press}
\end{align}

\subsection{Block Tumble}

\textbf{Observation Space:}
\begin{enumerate}
    \item 7-dimensional robot arm joint angles.
    \item The gripper width. This is assumed to be closed for this task.
    \item 6D pose of the end-effector.
    \item End-effector contact forces.
    \item 7D block pose with position in XYZ and orientation in quaternion.
    \item Block dimensions.
\end{enumerate}

\textbf{Reward Function:}
The reward function contains a distance improvement based metric that
rewards the agent to reach close to the object center.
The agent is also rewarded when the orientation of the block is improved towards the target orientation.
Formally, let $d_t$ be the euclidean distance between the end-effector and center of the block, $\theta_t$ be the absolute angle difference between the current pose of the block and the target pose.
The reward at timestep $t$ is then,
\begin{align}
    r_t &= 10\times\left(d_t-d_{t-1}\right) + 10\times\left(\theta_t - \theta_{t-1}\right) \\ \nonumber
    & \qquad + 100\times succ - 0.1 \label{eq:reward_block_tumble}.
\end{align}

\subsection{Door Open}

\textbf{Observation Space:}
\begin{enumerate}
    \item 7-dimensional robot arm joint angles.
    \item The scalar gripper width.
    \item 6D pose of the end-effector.
    \item End-effector contact forces.
    \item 3D door handle position, \emph{i.e.}, the position where the door handle meets the door frame.
    \item The door rotation angle, \emph{i.e.}, the angle by which the door is currently rotated. At $t=0$, this value is 0.
\end{enumerate}

\textbf{Reward Function:}
The reward function used consists of a distance based improvement reward for moving the end-effector towards the middle of the door handle. 
The agent is further rewarded for turning the door handle as well as opening the door.
There is also a time penalty, $0.1$ at each step, to allow the agent to complete the task as soon as possible.
There also exists a force penalty, $0.01$ at each step, to avoid the robot-arm hitting the door frame while trying to open the door.
Finally, there is a task reward for completing the task i.e. when the door has been opened beyond a threshold.
Formally, let $d_t$ be the euclidean distance between the end-effector and door handle.
Let $\theta_t$ by the rotation joint angle for the door angle and $\phi_t$ be the door frame rotation angle (both angles in radians).
The success of the task is defined as 
$succ = 1\left\{\phi_t > 0.436 \right\}$, \emph{i.e.}, if the door has been rotated more than $25^\circ$ degree.
The door opening task also has a locking mechanism which does not allow the door to be rotated unless $\theta_t > 1.22$, i.e. $70^\circ$ degree.
Thus, the reward function is defined as:
\begin{align}
    r_t &= 10\times\left(d_t-d_{t-1}\right) + 10\times\left(\theta_t - \theta_{t-1}\right) \\ \nonumber
    & \quad + 100\times\left(\phi_t-\phi_{t-1}\right) + 100\times succ - 0.1 - 0.01 \label{eq:reward_door_open}.
\end{align}

\section{Controller Details}
In the following sub-sections we detail the types of task-axes controllers used for each of the previous task.
As discussed in the main paper, 
each task-axes controller (TAC) is parameterized with some keypoint or axis.
For implementation details about each controller we refer the reader to \cite{sharma2020objaxes}.

\begin{table}[t]
\centering
\resizebox{\linewidth}{!}{%
\begin{tabular}{@{}llll@{}}
\toprule
Task & TAC (Manual) & TAC (Keypoints+Axes) & Keypoints \\ \midrule
Button Press & 5 & 14 & 2 \\
Block Tumble & 10 & 40 & 10 \\
Door Open & 8 & 51 & 4 \\ \bottomrule
\end{tabular}%
}
\caption{Controller Statistics for each of the tasks.}
\label{tab:controller_stats}
\end{table}

\subsection{Button Press}

For the button press task, we use the two different keypoints on the 
button object.
For each of these keypoints we define error-axis based position controllers. 
These controllers move towards the keypoint along the shortest possible axes and hence are only parameterized by keypoints and not axis.
Additionally, for each keypoint we also define position attractors.
Each of these attractor act along one particular object axis. 
This gives us a total of 6 controllers.
Additionally, we also define force controllers along each of the object
axis separately. 
We define force-controllers for both positive and negative axis direction. Thus, we get a total of 6 force controllers.
Overall, we have 14 controllers for this task.

\subsection{Block Tumble}

For the block tumble task, we use 10 different keypoints on the button object, distributed through the top surface of the object 
(Figure~\ref{fig:results_keypoint_params} middle).
Each of these keypoint helps the robot align the block to the target orientation in a slightly different manner.
This is because if the robot approaches the object from behind, 
using a keypoint towards the left edge of the block will orient the block towards the left edge,
while using a keypoint towards the right edge will orient it rightwards.
We use a large number of keypoints for this task to show that our approach is competitive compared to the baseline even in presence of a large numbre of controllers.
For each keypoint we use position attractors that act along the error axis and hence do not require an explicit axes.
This results in 10 controllers.
Additionally, for each keypoint we also use force-attractors that act along each of the object-axis direction separately.
For this task, force-attractors are more useful as compared to position-attractors since we require a minimum amount of force to tumble the block. 
This results in 30 controllers and 40 controllers overall.

\subsection{Door Open}
For the door open task we use 4 keypoints, each of which is labelled on the door handle (Figure~\ref{fig:results_keypoint_params} right).
For this task, in addition to position and force controllers we also use rotation as well as gripper controllers.
First, we use both open and close gripper controllers.
For each keypoint we also use error-axis based position attractors.
These result in 4 controllers (1 for each keypoint).
We also use handle rotation controllers, which uniformly select one of the EE-axis and one of the handle axes and aims to orient the EE-axis parallel to the handle axis.
For these handle rotation controllers, we choose both positive and negative axes direction. 
This results in a set of 18 controllers.
We also use curl-attractors for this task, 
which allow the robot end-effector to rotate around the door handle. These controllers are particularly useful when the robot fails to grasp the door handle because of slippage or improper grasps.
We define one curl attractor for each position and along each axis of the door handle. This yields 12 controllers.
We also define force attractors that act indepdnent of any position and apply force in the negative direction of the handle axes.
For each keypoint we also have force attractors that act along each of the axis, resulting in 12 controllers.
This results in a total of 51 controllers for this task.

\section{Training Details}
\label{app:train_details}

\subsection{PPO Hyperparameters}
\begin{table}[!h]
\centering
\begin{tabular}{l|lll}
\toprule
        Parameters         & Button Press & Block Tumble & Door-Open    \\
\midrule
num steps                  & $120$               & $120$               & $240$                              \\
discount factor            & $0.995$             & $0.995$             & $0.995$                         \\
entropy coefficient        & $0.01$              & $0.01$              & $[0.01, 0.1]$                     \\
learning rate              & $2.5\times 10^{-4}$ & $2.5\times 10^{-4}$ & $2.5\times 10^{-4}$  \\
value loss coefficient     & $0.5$               & $0.1$              & $0.5$                              \\
max gradient norm          & $0.5$               & $0.5$               & $0.5$                             \\
lambda                     & $0.95$              & $0.95$              & $0.95$                           \\
num minibatches            & $30$                & $30$                & $40$                               \\
num opt epochs             & $4$                 & $4$                 & $4$                                 \\
clip range                 & $0.2$               & $0.2$               & $0.2$                           \\
\bottomrule
\end{tabular}
\caption{PPO Hyperparameters Across All Tasks.}
\label{tab:hparams_ppo}
\end{table}

Table~\ref{tab:hparams_ppo} lists the hyperparameters for PPO used for each of the experiments, to train the task policy.
For each environment the episode length is $120$. Also, we run $\approx 20$ environments in parallel to collect data during PPO training.

\subsection{Reinforce Hyperparameters}
We use Reinforce \cite{williams1992simple} to train the keypoint selection policy, using the learned task policy to evaluate the proposed keypoints.
The architecture for keypoint prediction model is presented in Section~\ref{app:keypoint-model}.
To train the keypoint model, we use the Adam optimizer \cite{kingmaadam}. We set an initial learning rate of $1e-3$, which is decayed by $0.3$ after 200 and 1000 iterations respectively. 
At each iteration, we collect keypoint predictions from $24$ environments simultaneously.
We update the model based on the policy gradient loss, which is averaged over all environments.

\begin{figure}[t]
    \centering
    \includegraphics[width=0.96\linewidth]{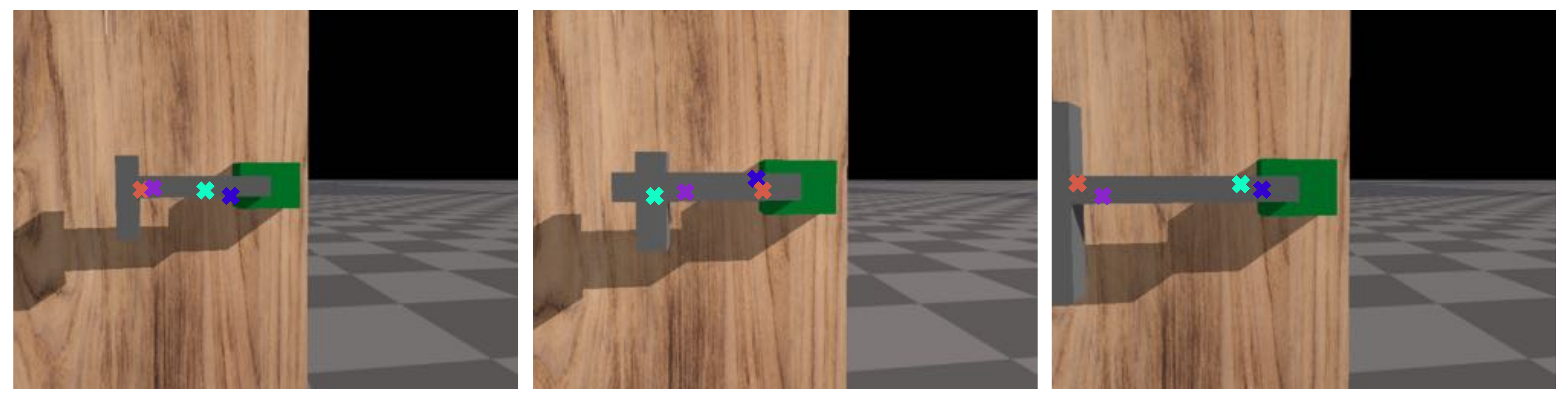}
    \caption{Keypoint Parameters inferred using learned dense object network $\phi$ for Config-A door handles. 
    We use the same set of reference pixels (keypoints) throughout for all configs (see Figure~\ref{fig:results_keypoint_params} for reference).
    }
    \label{fig:results_dense_correspondence_fail}
\end{figure}

\subsection{Learning Keypoint Parameters: Model Details}
\label{app:keypoint-model}

We learn the keypoint controller parameters from visual input using a convolutional neural network architecture.
More precisely, the input to our model is an RGB image with dimensions $(128, 128)$.
We initially use a series of VGG style convolutional layers, each with $64, 128, 256, 512$ channels respectively. 
Each convolutional layer has a kernel of size $3x3$ and padding of size $1$, which is followed by the ReLU non-linearity.
Also, each convolutional layer is followed by a MaxPool layer with kernel of size $2$ and stride $2$.
The output of this series of convolutional layers is passed through another convolutional layer which downsamples the number
of channels from $512$ to $32$, using a kernel of size $2$ and padding of size $2$.
The output from this is an $8x8$ image with 32 channels.

This output is upsampled through two UnetBlocks \cite{ronneberger2015u}, each outputting 16 and 4 channels respectively.
Finally, we use a convolutional layer that downsamples the 4 channels to 1 channel using a kernel of size 1, which is subsequently 
passed through a softmax.
Thus, our output is a $32\times 32$ image, where each pixel represents the probability value of choosing this pixel as the keypoint.
Since each pixel in the output image corresponds to a $4x4$ pixel range in the original input, we refer to the output pixels as superpixels.

To find the 3D location from the selected superpixel, we sample all pixels within this superpixel and use the depth values at each pixel to 
find their corresponding 3D positiosn. We average these 3D positions to find the 3D location for the selected superpixel.

block-tumble-refine-train-results.png

\begin{figure}[t]
    \centering
    \includegraphics[width=0.98\linewidth]{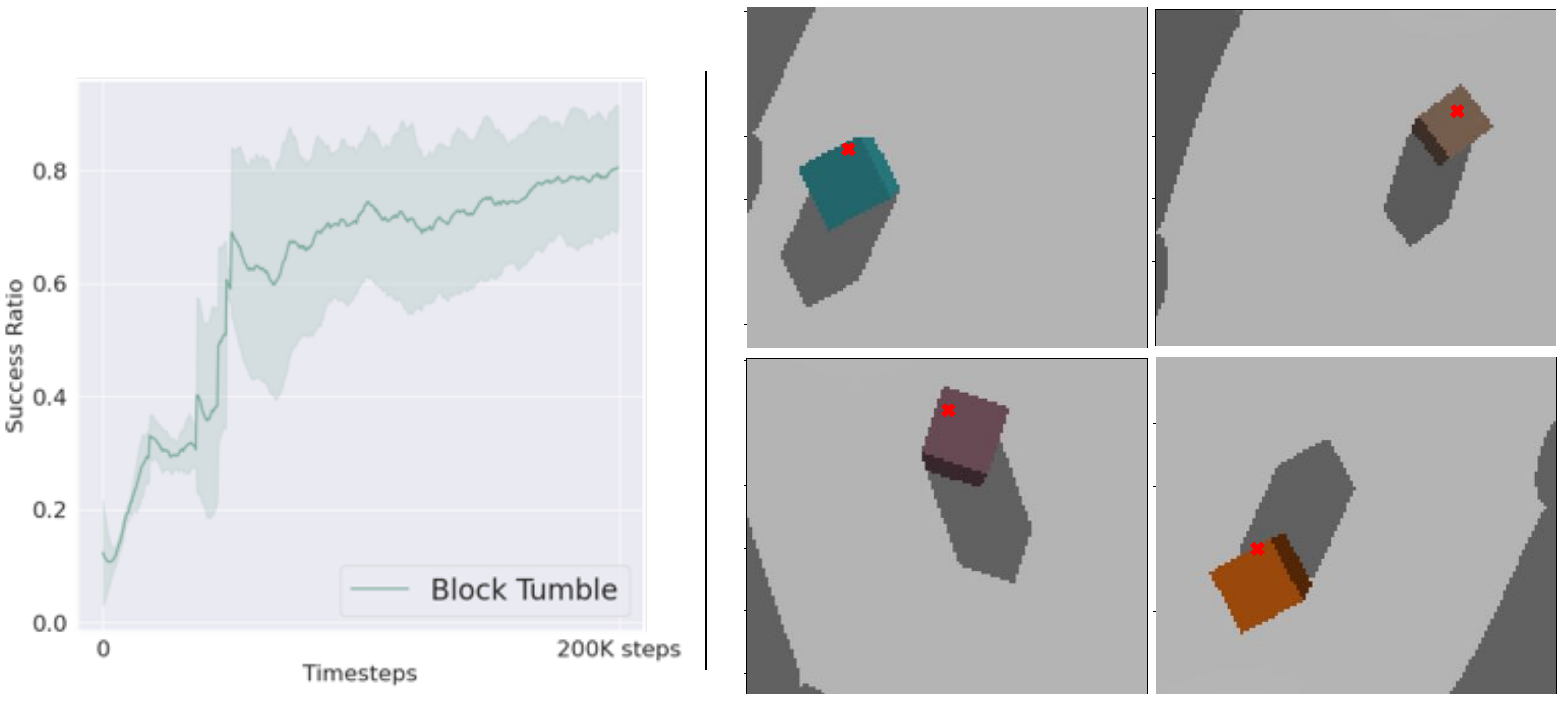}
    \caption{Keypoint parameters learned from raw input image using only a learned block tumble task policy. See subsection~\ref{subsec:block-tumble-refine-keypoint} for details.}
    \label{fig:results_block_tumble_keypoints}
\end{figure}

\section{Additional Results}
\label{app:results}
In this section we show some additional results related with the 
tasks and experiments in the main paper.
First, we show that dense object nets can fail to find semantically relevant keypoints when used on objects that are not quite similar with the objects used to train the descriptors.
Figure~\ref{fig:results_dense_correspondence_fail} shows the keypoints inferred by the trained dense object network $\phi$ on Config-A door handles (see Figure~\ref{fig:results_refine_keypoints_1} for reference). 
We note that we use the same set of reference pixels (reference keypoints) as in the main paper (see Figure~\ref{fig:results_keypoint_params}.
However, given that the object shapes for both set of objects is 
slightly different, 
the learned dense object descriptors fail to generalize to such different objects.
This is not unexpected and shows a limitation of solely using visual correspondence for generalization.

\subsection{Learning Controller Parameters with Learned Task Policy - Block Tumble}
\label{subsec:block-tumble-refine-keypoint}
We also show results for learning position controller targets for the Block Tumble task using the learned task policy.
Although previously we used multiple keypoints to learn the block tumble task policy (see Section~\ref{sec:results} for discussion), 
most of the used keypoints are not used by the task policy. 
This is because to perform the task, the agent only needs to reach close to some point on the block and flip it. 
This can be achieved by using one relevant keypoint. 

We use this insight to learn a relevant keypoint parameter from raw visual data for the block tumble task. 
As before we get the task policy by bootstrapping keypoint parameters using dense correspondence learning. However, this task policy only uses 1 keypoint. 
Also, as discussed in \ref{app:keypoint-model} instead of directly predicting the keypoint pixel, we predict the relevant superpixel of size $4\times4$. 
Figure~\ref{fig:results_block_tumble_keypoints} (Left) shows the training plot for learning the relevant keypoint parameter using reinforce.
Since the task setup only involves a object on the block,
the input image for the keypoint model is not very complex and hence the network is able to find the relevant keypoint on or sometimes near the object. 
Figure~\ref{fig:results_block_tumble_keypoints} (Right) shows some qualitative results for the keypoints inferred on the block. 
As seen in the above image, most keypoints are found on the object (often near to the edge). 




%% file: main.bbl
\begin{thebibliography}{10}
\providecommand{\url}[1]{#1}
\csname url@samestyle\endcsname
\providecommand{\newblock}{\relax}
\providecommand{\bibinfo}[2]{#2}
\providecommand{\BIBentrySTDinterwordspacing}{\spaceskip=0pt\relax}
\providecommand{\BIBentryALTinterwordstretchfactor}{4}
\providecommand{\BIBentryALTinterwordspacing}{\spaceskip=\fontdimen2\font plus
\BIBentryALTinterwordstretchfactor\fontdimen3\font minus
  \fontdimen4\font\relax}
\providecommand{\BIBforeignlanguage}[2]{{%
\expandafter\ifx\csname l@#1\endcsname\relax
\typeout{** WARNING: IEEEtran.bst: No hyphenation pattern has been}%
\typeout{** loaded for the language `#1'. Using the pattern for}%
\typeout{** the default language instead.}%
\else
\language=\csname l@#1\endcsname
\fi
#2}}
\providecommand{\BIBdecl}{\relax}
\BIBdecl

\bibitem{manuelli2019kpam}
L.~Manuelli, W.~Gao, P.~Florence, and R.~Tedrake, ``kpam: Keypoint affordances
  for category-level robotic manipulation,'' \emph{arXiv preprint
  arXiv:1903.06684}, 2019.

\bibitem{florence2018dense}
P.~R. Florence, L.~Manuelli, and R.~Tedrake, ``Dense object nets: Learning
  dense visual object descriptors by and for robotic manipulation,'' in
  \emph{Conference on Robot Learning}, 2018, pp. 373--385.

\bibitem{ganapathi2020learning}
A.~Ganapathi, P.~Sundaresan, B.~Thananjeyan, A.~Balakrishna, D.~Seita,
  J.~Grannen, M.~Hwang, R.~Hoque, J.~E. Gonzalez, N.~Jamali \emph{et~al.},
  ``Learning to smooth and fold real fabric using dense object descriptors
  trained on synthetic color images,'' \emph{arXiv preprint arXiv:2003.12698},
  2020.

\bibitem{sharma2020objaxes}
M.~Sharma, J.~Liang, J.~Zhao, A.~LaGrassa, and O.~Kroemer, ``Learning to
  compose hierarchical object-centric controllers for robotic manipulation,''
  \emph{arXiv preprint arXiv:2011.04627}, 2020.

\bibitem{mason1981compliance}
M.~T. Mason, ``Compliance and force control for computer controlled
  manipulators,'' \emph{IEEE Transactions on Systems, Man, and Cybernetics},
  vol.~11, no.~6, pp. 418--432, 1981.

\bibitem{raibert1981hybrid}
M.~H. Raibert and J.~J. Craig, ``Hybrid position/force control of
  manipulators,'' 1981.

\bibitem{ballard1984task}
D.~H. Ballard, ``Task frames in robot manipulation.'' in \emph{AAAI}, vol.~19,
  1984, p. 109.

\bibitem{muhlig2009automatic}
M.~M{\"u}hlig, M.~Gienger, J.~J. Steil, and C.~Goerick, ``Automatic selection
  of task spaces for imitation learning,'' in \emph{2009 IEEE/RSJ international
  conference on intelligent robots and systems}.\hskip 1em plus 0.5em minus
  0.4em\relax IEEE, 2009, pp. 4996--5002.

\bibitem{berenson2011task}
D.~Berenson, S.~Srinivasa, and J.~Kuffner, ``Task space regions: A framework
  for pose-constrained manipulation planning,'' \emph{The International Journal
  of Robotics Research}, vol.~30, no.~12, pp. 1435--1460, 2011.

\bibitem{king2016rearrangement}
J.~E. King, M.~Cognetti, and S.~S. Srinivasa, ``Rearrangement planning using
  object-centric and robot-centric action spaces,'' in \emph{2016 IEEE
  International Conference on Robotics and Automation (ICRA)}.\hskip 1em plus
  0.5em minus 0.4em\relax IEEE, 2016, pp. 3940--3947.

\bibitem{kober2015learning}
J.~Kober, M.~Gienger, and J.~J. Steil, ``Learning movement primitives for force
  interaction tasks,'' in \emph{2015 IEEE International Conference on Robotics
  and Automation (ICRA)}.\hskip 1em plus 0.5em minus 0.4em\relax IEEE, 2015,
  pp. 3192--3199.

\bibitem{ureche2015task}
A.~L.~P. Ureche, K.~Umezawa, Y.~Nakamura, and A.~Billard, ``Task
  parameterization using continuous constraints extracted from human
  demonstrations,'' \emph{IEEE Transactions on Robotics}, vol.~31, no.~6, pp.
  1458--1471, 2015.

\bibitem{migimatsu2020object}
T.~Migimatsu and J.~Bohg, ``Object-centric task and motion planning in dynamic
  environments,'' \emph{IEEE Robotics and Automation Letters}, vol.~5, no.~2,
  pp. 844--851, 2020.

\bibitem{manschitz2020learning}
S.~Manschitz, M.~Gienger, J.~Kober, and J.~Peters, ``Learning sequential force
  interaction skills,'' \emph{Robotics}, vol.~9, no.~2, p.~45, 2020.

\bibitem{peternel2017method}
L.~Peternel, L.~Rozo, D.~Caldwell, and A.~Ajoudani, ``A method for derivation
  of robot task-frame control authority from repeated sensory observations,''
  \emph{IEEE Robotics and Automation Letters}, vol.~2, no.~2, pp. 719--726,
  2017.

\bibitem{conkey2019learning}
A.~Conkey and T.~Hermans, ``Learning task constraints from demonstration for
  hybrid force/position control,'' in \emph{2019 IEEE-RAS 19th International
  Conference on Humanoid Robots (Humanoids)}.\hskip 1em plus 0.5em minus
  0.4em\relax IEEE, 2019, pp. 162--169.

\bibitem{florence2019self}
P.~Florence, L.~Manuelli, and R.~Tedrake, ``Self-supervised correspondence in
  visuomotor policy learning,'' \emph{IEEE Robotics and Automation Letters},
  vol.~5, no.~2, pp. 492--499, 2019.

\bibitem{Gao_kpam2}
W.~{Gao} and R.~{Tedrake}, ``kpam 2.0: Feedback control for category-level
  robotic manipulation,'' \emph{IEEE Robotics and Automation Letters}, pp.
  1--1, 2021.

\bibitem{qin2019keto}
Z.~Qin, K.~Fang, Y.~Zhu, L.~Fei-Fei, and S.~Savarese, ``Keto: Learning keypoint
  representations for tool manipulation,'' 2019.

\bibitem{schmidt2016self}
T.~Schmidt, R.~Newcombe, and D.~Fox, ``Self-supervised visual descriptor
  learning for dense correspondence,'' \emph{IEEE Robotics and Automation
  Letters}, vol.~2, no.~2, pp. 420--427, 2016.

\bibitem{masson2016reinforcement}
W.~Masson, P.~Ranchod, and G.~Konidaris, ``Reinforcement learning with
  parameterized actions,'' in \emph{Proceedings of the AAAI Conference on
  Artificial Intelligence}, vol.~30, no.~1, 2016.

\bibitem{hausknecht2015deep}
M.~Hausknecht and P.~Stone, ``Deep reinforcement learning in parameterized
  action space,'' \emph{arXiv preprint arXiv:1511.04143}, 2015.

\bibitem{williams1992simple}
R.~J. Williams, ``Simple statistical gradient-following algorithms for
  connectionist reinforcement learning,'' \emph{Machine learning}, vol.~8, no.
  3-4, pp. 229--256, 1992.

\bibitem{schulman2017proximal}
J.~Schulman, F.~Wolski, P.~Dhariwal, A.~Radford, and O.~Klimov, ``Proximal
  policy optimization algorithms,'' \emph{arXiv preprint arXiv:1707.06347},
  2017.

\bibitem{stablebaselines}
A.~Hill, A.~Raffin, M.~Ernestus, A.~Gleave, A.~Kanervisto, R.~Traore,
  P.~Dhariwal, C.~Hesse, O.~Klimov, A.~Nichol, M.~Plappert, A.~Radford,
  J.~Schulman, S.~Sidor, and Y.~Wu, ``Stable baselines,''
  \url{https://github.com/hill-a/stable-baselines}, 2018.

\bibitem{kingmaadam}
\BIBentryALTinterwordspacing
D.~P. Kingma and J.~Ba, ``Adam: A method for stochastic optimization,'' in
  \emph{ICLR (Poster)}, 2015. [Online]. Available:
  \url{http://arxiv.org/abs/1412.6980}
\BIBentrySTDinterwordspacing

\bibitem{ronneberger2015u}
O.~Ronneberger, P.~Fischer, and T.~Brox, ``U-net: Convolutional networks for
  biomedical image segmentation,'' in \emph{International Conference on Medical
  image computing and computer-assisted intervention}.\hskip 1em plus 0.5em
  minus 0.4em\relax Springer, 2015, pp. 234--241.

\end{thebibliography}
